\documentclass{article}

\usepackage{times}
\usepackage{graphicx} 
\usepackage{subfigure}

\usepackage{natbib}

\usepackage{algorithm}
\usepackage{algorithmic}
\usepackage{amssymb}
\usepackage{amsthm}
\usepackage{amsmath}
\usepackage{url}

\usepackage{hyperref}

\usepackage[accepted]{icml2014}

\icmltitlerunning{Submission and Formatting Instructions for uLearnBio 2014}

\begin{document}

\twocolumn[
\icmltitle{Unsupervised Pretraining Encourages Moderate-Sparseness}

\icmlauthor{Jun Li, Wei Luo, and Jian Yang}{junl.njust@gmail.com;cswluo@gmail.com;csjyang@njust.edu.cn}
\icmladdress{School of Computer Science and Technology, Nanjing University of Science and Technology, Nanjing, China, 210094}
\icmlauthor{Xiaotong Yuan}{xtyuan1980@gmail.com}
\icmladdress{School of Information and Control, Nanjing University of Information Science and Technology, Nanjing, China, 210044}

\icmlkeywords{boring formatting information, machine learning, ICML}

\vskip 0.3in
]

\begin{abstract}
It is well known that direct training of deep neural networks will generally  lead to poor results.
  A major progress in recent years is the invention of various pretraining methods to initialize network parameters and it was shown that such methods lead to good prediction performance. However, the reason for the success of pretraining has not been fully understood, although it was argued that regularization and better optimization play certain roles. This paper provides another explanation for the effectiveness of pretraining, where we show pretraining leads to a sparseness of hidden unit activation in the resulting neural networks. The main reason is that the pretraining models can be interpreted as an adaptive sparse coding. Compared to deep neural network with sigmoid function, our experimental results on MNIST and Birdsong further support this sparseness observation.
\end{abstract}

\section{Introduction}
Deep neural networks (DNNs) have found many successful applications in recent years.
However, it is well-known that if one trains such networks with the standard back-propagation algorithm from randomly initialized parameters, one typically ends up with models that have poor prediction performance.
A major progress in DNNs research is the invention of pretraining techniques for deep learning \cite{Hinton2006a,Hinton2006b,Bengio2006,Bengio2009a,Bengio2012}. The main strategy is to employ layer-wise unsupervised learning procedures to initialize the DNN model parameters. A number of such unsupervised training techniques have been proposed, such as restricted Boltzmann machines (RBMs) in \cite{Hinton2006a}, and denoising autoencoders (DAEs) in \cite{Vincent2008}.
Although these methods show strong empirical performance, the reason of their success has not been fully understood.

Two reasons were offered in the literature to explain the advantages of unsupervised learning procedure \cite{Erhan2010,Larochelle2009}: the regularization effect and the optimization effect. The regularization effect says that pretraining provides regularization which initialize the parameters in the basin of attraction to a ``good'' local minimum. The optimization effect says that the pretraining leads to better optimization so that the initial value is close to a local minimum with a lower objective value than that can be achieved with random initialization. Based on experimental evidences, some researchers confirm that the pretraining can learn invariant representations and selective units \cite{Goodfellow2009}.

\textbf{Our Contributions:} We study why the pretraining encourages moderate-sparseness. The main reason is that the pretraining models can be interpreted as an adaptive sparse coding. This coding is approximated by a sparse encoder, which is implemented by adaptively filtering out a lot of features that are not present in the input and suppressing the responses of some features that are not significant in the input. We further conduct experiments to demonstrate that it is a sparse regularization (the hidden units become more sparsely activated).

\section{Previous Works}
In this part we review some advantages of the pretraining methods.

Distributed representations and deep architectures play an important role in deep learning methods. A distributed representation (an old idea) can capture a very great number of possible input configurations \cite{Bengio2009a}. Deep architectures can promote the re-use of features and lead to abstract more invariant features for most local changes of the inputs \cite{Bengio2012}. However, it is hard to use the \emph{Back-Propagation} to train DNNs with two traditional activation functions (the sigmoid function $1/(1 + e^{-x})$ and the hyperbolic tangent $\tanh(x)$). Luckily, \cite{Hinton2006a} proposes a unsupervised pretraining method to initialize the DNNs model parameters and learn good representations. The regularization effect and the optimization effect are used to explain the main advantages of the pretraining method \cite{Erhan2010,Larochelle2009}.

To better understand what the pretraining models learn in deep architectures, \cite{Goodfellow2009} find that the pretraining methods can learn invariant representations and selective units. Some researchers use the linear combination of previous units \cite{Lee2009} and the maximizing activation \cite{Erhan2010} to visualize the feature detectors (or invariance manifolds or filters) in an arbitrary layer. Fig. 1 of \cite{Erhan2010} and Fig. 3 of \cite{Lee2009} show that the first, second and third layer can learn edge detectors, object parts, and objects respectively. Based the distributed and invariant representations, \cite{Larochelle2009,Bengio2012,Bengio2013} further confirm that the pretraining methods tend to do a better job at disentangling the underlying factors of variation, such as objects or object parts.

Compared to DNNs with sigmoid function, we confirm that the pretraining methods encourage moderate-sparseness as the detectors filter out a lot of features that are not present in the input. In general, there is an illusion that unsupervised pretraining methods tend to learn non-sparse representations because it does not meet the conventional sparse methods \cite{Zhangl2011,Yang2012,Yang2013}. The conventional methods consider the idea of introducing a form of sparsity regularization. Most ways have been proposed by directly penalizing the outputs of hidden units, such as $L_1$ penalty, $L_1/L_2$ penalty and Student-t penalty. But, the pretraining methods implement sparseness by filtering out a lot of irrelevant features.

\section{Pretraining Model}

There is a classic pretraining models: RBMs. An RBMs is an energy-based generative model defined over a visible layer and a hidden layer. The visible layer is fully connected to the hidden layer via symmetric weights $W$, while there have no connections between units of the same layer. The number of visible units $x$ and hidden units $h$ are denoted by $d_x$ and $d_h$, respectively. Additionally, visible units and hidden units receive input from bias - $c$ and $b$ respectively. The energy function is denoted by $\eta(x,h)$:
\begin{align}
\eta(x,h)=-h^{\textbf{T}}Wx-c^{\textbf{T}}x-b^{\textbf{T}}h
\label{eq:ef}
\end{align}

The probability that the network assigns to visible units $x$ is
\begin{align}
p(x)= \frac{1}{Z} \sum_{h}e^{-\eta(x,h)} \qquad  Z=\sum_{x,h}e^{-\eta(,h)}
\end{align}
where $Z$ is the partition function or normalizing constant. Because there are no direct connections between hidden (or visible) units, it is very easy to sample from the conditional functions taking the form:
\begin{align}
p(x|h)=\prod_{i=1}^{d_x}p(x_i|h) \qquad p(h|x)=\prod_{j=1}^{d_h}p(h_j|x)
\label{eq:prod}
\end{align}
where $p(h_j=1|x)=f\left(\sum_{i=1}^{d_x}W_{ji}x_i+b_j\right)$, $p(x_i=1|h)=f\left(\sum_{j=1}^{d_h}W_{ji}h_j+c_i\right)$ and $f(t)$ is a logistic sigmoid functions: $f(t) = 1/(1 + \text{exp}(-t))$.

The training is to use $CD-1$ algorithm \cite{Hinton2002} to minimize the likelihood of the data: $-\log p(x)$.

\section{Unsupervised Pretraining Encourages Moderate-Sparseness}
\label{sec:whysparse}
In this section, we denote that the sparse regularization with more overlapping groups in low layer or less in high layer is called the Moderate-Sparseness. We mainly consider the multi-class problem to explain why the unsupervised pretraining encourages moderate-sparseness since DNNs with the pretraining has been used to achieve state-of-the-art results on classification tasks. There are two reasons. First, we show a new viewing that the pretraining model is an adaptive sparse coding. Second, because the pretraining can train the "good" feature detectors, we discuss that how the feature detectors can lead to moderate-sparseness. Finally, we measure the moderate-sparseness.

To start off the discussion, there are two natural assumptions to $m$-class training set. \emph{Assumption 1:} Every class has a balanced number of samples and there are a lot of common raw features (pixels or MFCC features) among samples of the same class \cite{Zhangc2012}. \emph{Assumption 2:} There are some similar raw features among samples of different classes since they share some common ones \cite{Amit2007}.

\subsection{A New Viewing of Pretraining Model}

Pretraining Model (such as RBMs) is an adaptive sparse coding. The explanation is as follow. By the results of \cite{Bengio2009b}, the pretraining model (RBMs training is to minimize $-\log p(x)$) is also approximated by minimizing a reconstruction error criterion:
\begin{align}
-(\log p(x|\widehat{h})+\log p(h|\widehat{x}))
\label{eq:rbmre1}
\end{align}
where $E_h[p(h|x)]$ is the mean-field output of the hidden units given the observed input $x$ and $E_x[p(x|h)]$ is the mean-field output of the visible units given the representation $h$ sampled from $p(h_j=1|x)=f\left(\sum_{i=1}^{d_x}W_{ji}x_i+b_j\right)$. The $-\log p(x|\widehat{h})$ and $-\log p(h|\widehat{x})$ can be regard as a decoder and an encoder, respectively.

From the second parts of \eqref{eq:prod} and \eqref{eq:rbmre1}, every hidden unit $p(h_j=1|x)=f\left(\sum_{i=1}^{d_x}W_{ji}\widehat{x}_i+b_j\right),(j=1,\cdots,d_h)$ can be further interpreted as a feature detector (or invariance manifolds or filters) because the hidden unit is active (or non-active), that means, the detector should respond strongly (or weakly) when the corresponding feature is present (or absent) in the input \cite{Goodfellow2009}.

Amazedly the pretraining can train edge feature detectors in low layer and objects (or object parts) in high layer \cite{Lee2009,Larochelle2009,Bengio2012}. \emph{Given an input, the feature detectors naturally filter out a lot of features that are not present in the input and suppress the responses of some features that are not significant in the input.} Clearly, those detectors result in sparseness.

\textbf{Relationship with sparse coding}: Sparse coding is to find the dictionary $D$ and the sparse representation $h$ to minimize the most popular form:
\begin{align}
  \|x-Dh\|^2 + \lambda_{sc}\| h\|_1
\label{eq:sc}
\end{align}
where $\lambda_{sc}$ also is a hyper-parameter. Obviously, the first part of RBMs \eqref{eq:rbms} is similar to the first part of sparse coding \eqref{eq:sc} as they are decoders. The sparse coding is directly to penalize the $L_1$ norm of the hidden representation $h$. But in RBMs \eqref{eq:rbms} the $h$ is approximated by the sparse encoders (feature detectors), which filter out a lot of irrelevant features. In next subsection we shall discuss that how the feature detectors can lead to moderate-sparseness.

\subsection{Lead To Moderate-Sparseness}

\textbf{Low-layer}: Based on the assumptions the pretraining models averagely distribute all edge feature detectors to the hidden units in low layer as every class has a same number of samples. Assumption 1 shows that every class has a same number of edge feature detectors and there are a lot of common edge ones in the same class. Clearly, the edge feature detectors find out the edge features belonged to self-class, suppress the responses of some nonsignificant edge features, and filter out a lot of edge features related to the other classes. Suppose that there are $N$ hidden units (edge feature detectors) and a $m$-class dataset, every class ideally has $\frac{N}{m}$ ones. Given input samples of a class, thus, the $\frac{N}{m}$ hidden units belonged to the class are activated or weakly responded and the remaining $N-\frac{N}{m}$ units are not activated (corresponding sparseness that is measured by \eqref{eq:hspm}). Moreover, there are the common activation units (corresponding group).

Simultaneously, assumption 2 shows that there are some similar edge feature detectors among different classes. Different classes share some edge feature detectors corresponded to hidden units, which are also activated. The activated units results in more overlapping activation units in low layer. The activation overlapping degree is measured by \eqref{eq:aod}. Combined with regularization effect \cite{Erhan2010}, therefore, we obtain the first result (A1) that \emph{the unsupervised pretraining is a sparse regularization with more-overlapping groups in low layer}.

\textbf{High-layer}: In high layer the pretraining goes on to train object or object part features detectors from the edge features. Similarly to the analysis in low layer, the hidden units are more sparsely activated or weakly responded in high layer. Moreover, the activation overlapping degree is lower than one in low layer because the pretraining can potentially lead to do a better job at disentangling the objects or object parts \cite{Larochelle2007,Bengio2013}. Thus, we obtain the second result (A2) that \emph{the unsupervised pretraining is a sparse regularization with less (or no)-overlapping groups in high layer}.

In DNNs without the pretraining the most hidden units of every layer are always activated and correspond to terrible feature detectors, which are the important causes of difficult classification problems. For classification tasks, it is always desirable to extract features that are most effective for preserving class separability \cite{Wong2011} and collaboratively representing objects \cite{Zhangl2011,Yangm2012}. The pretraining firmly grasps the those benefits. The more activation overlapping units can capture the collaborative features in low layer and the less or no activation overlapping units can capture the separability in high layer.

\subsection{Sparseness Measure}

For better understanding the pretraining, we tried to find sparseness, more-overlapping and no-overlapping characteristics of DNNs with or without the pretraining. So, Hoyer's sparseness measure and activation overlapping degree are defined as followings.

The Hoyer's sparseness measure (HSPM) \cite{Hoyer2004} is based on the relationship between the $L_1$ norm and the $L_2$ norm. The HSPM of a $n$ dimensional vector $h$ is defined as follows:
\begin{equation}
HSPM(h)=\frac{\sqrt{n}-(\sum_{i=1}^n|h_i|)/\sqrt{\sum_{i=1}^n h_i^2}}{\sqrt{n}-1},
\label{eq:hspm}
\end{equation}
This measure has good properties, which is in the interval $[0, 1]$ and on a normalized scale. It's value more close to $1$ means that there are more zero components in the vector $h$. We denote $|\cdot|$ absolute value of a real number and give the following definitions about AOD.

\textbf{Definition 1:} A hidden unit $i$ is said to be active if the absolute value of its activation $h_i$ is above a threshold $\tau$, that is $|h_i|>\tau$. And a hidden unit $i$ is called un-active if $|h_i|<\tau$.

\textbf{Definition 2:} A vector $z$ is said to be an activation binary-vector of a $d$ dimensional representation $h$ if some representation units are active when the corresponding features are present in $x$, and otherwise are not active when they are absent. Formally, the activation binary-vector $z=z(h)$ is defined as:
\begin{equation}
 z_i=z_i(h_i)=\left\{
   \begin{array}{ll}
   1, & \hbox{$|h_{i}|\geq\tau$;} \\
   0, & \hbox{$|h_{i}|< \tau$.}
   \end{array}
   \right.
 ; i=1,\cdots,d_h
\label{eq:selbinunit}
\end{equation}
where $\tau$ is a threshold.

To indicate the present feature in the input, we select a threshold $\tau$ that does not change the reconstruction data, that is $\|f(W^Ts+c)-f(W^Th+c)\|^2<0.05$\footnote{We select $0.05$ because it is small enough.}, where $h=f(Wx+b)$ and the vector $s=s(x)$ of a sample $x$ is defined as:
\begin{equation}
 s_i=\left\{
   \begin{array}{ll}
   h_{i}, & \hbox{$|h_{i}|\geq\tau$;} \\
   0, & \hbox{$|h_{i}|< \tau$.}
   \end{array}
   \right.
 ; i=1,\cdots,d_h
\label{eq:selunit}
\end{equation}

\textbf{Definition 3:} An activation binary-vector $Z(\mathcal{X})$ of a sample set $\mathcal{X}$ is an activation binary-vector of the mean value $\overline{x}$ among all samples $x (x\in\mathcal{X})$. It is defined as:
\begin{equation}
Z(\mathcal{X})= z(\overline{x}) \quad \overline{x}= \frac{\sum_{x\in\mathcal{X}}x}{m}
\label{eq:selclassunit}
\end{equation}
where $z(\overline{x})$ is defined in \eqref{eq:selbinunit} and $m$ is the number of sample in the set $\mathcal{X}$.
The activation overlapping degree (AOD) simply calculates the percentage of activation unites that are simultaneously selected by different classes $\mathcal{X}_i (i=1,\cdots,m)$. AOD among a set $H$ is defined as:
\begin{equation}
AOD(\mathcal{X}_1,\cdots,\mathcal{X}_m)= \frac{\sum_{j=1}^n z_j}{n}\quad z=\bigwedge_{i=1}^m Z(\mathcal{X}_i)
\label{eq:aod}
\end{equation}
where $H=\{\mathcal{X}_1,\cdots,\mathcal{X}_m\}$, $z$ is a binary-vector that is a logical conjunction on all activation binary-vectors $Z(\mathcal{X}_i),i=1,\cdots,m$ and $Z(\mathcal{X}_i)$ is defined in \eqref{eq:selclassunit}.

AOD, which is in the interval $[0, 1]$, is used to measure the percentage of activation overlapping units in different classes. It's value more close to $0$ means that there are few activation overlapping units and it is easier to separate the different classes.

\section{Experiments}
\label{sec:exp}

In this section, we use deep neural networks to do experiments. A standard architecture for DNNs consists of multiple layers of units in a directed graph, with each layer fully connected to the next one. The nodes of the inter-layers are called hidden units. Each hidden unit is passed through a standard sigmoid functions. The objective of learning is to find the optimal network parameters so that the network output matches the target closely. The output can be compared to a target vector through a squared loss function or an negative log-likelihood loss function. We employ the standard {\em back-propagation} algorithm to train the model parameters (the connected weights) \citep{Bishop2006}.

We denote that Dsigm: DNNs with standard sigmoid functions, DpRBMs: DNNs only pretrained with RBMs and DBNs: deep belief networks pretrained with RBMs and finely tuned.

\textbf{Datasets}: We present experimental results on standard benchmark datasets: MNIST\footnote{http://yann.lecun.com/exdb/mnist/} and Birdsong\footnote{http://sabiod.univ-tln.fr/icml2013/BIRD-SAMPLES/}
The pixel intensities of all datasets are normalized to $[0,1]$. \textbf{MNIST} dataset has 60,000 training samples and 10,000 test samples with $28\times28$ pixel greyscale images of handwritten digits 0-9. \textbf{Birdsong}\footnote{TRAIN SET has 30 sec $\times$ 35 bird recordings and TEST SET has 150sec $\times$ 3 mics $\times$ 90 recordings. There are not labels in TEST SET. So we divide the TRAIN SET to a new train set and a new test set(We randomly select 3,000 train samples with 16 MFCC features, the rest are test samples in every recording.).} dataset has 70,000 training samples and 200,690 test samples with 16 MFCC features.

To speed-up training, we subdivide training sets into mini-batches, each containing 100 cases, and the model parameter is updated after each minibatch using the averages. Weights are initialized with small random values sampled from a normal distribution with zero mean and standard deviation of $0.01$. Biases are initialized with zeros. For simplicity, we use a constant learning rate chosen from $\{1, 0.1, 0.05, 0.01\}$. Momentum is also used to speed up learning. The momentum starts at a value of $0.5$ and linearly increases to $0.90$ over half epochs, and stays at $0.9$ thereafter. The $L_2$ regularization parameter for the weights is fixed at $0.0001$.

\begin{table}[!t]
\renewcommand{\arraystretch}{1.3}
\caption{Hoyer's sparseness measures (HSPM) of DNNs (500-500-2000) on MNIST.}
\label{tab:hspm552}
\vskip -0in
\centering
\small{
\begin{tabular}{c|c|c|c|c|c}
\hline\hline
         & dataset & 1st  & 2nd  & 3rd  & error  \\
\hline
DBNs   &  0.63   & 0.39 & 0.53 & 0.63 & 1.17$\%$  \\
DpRBMs  &  0.63   & 0.39 & 0.58 & 0.67 &  \\
Dsigm  & 0.63    & 0.17 & 0.18 & 0.06 & 2.01$\%$ \\
\hline\hline
\end{tabular}}
\vskip -0.3in
\end{table}

\subsection{Sparseness Comparison}
\label{sec:sparsecomp}

\begin{figure*}[ht]
\vskip -0.1in
\begin{center}
\centerline{\includegraphics[width=1.5\columnwidth]{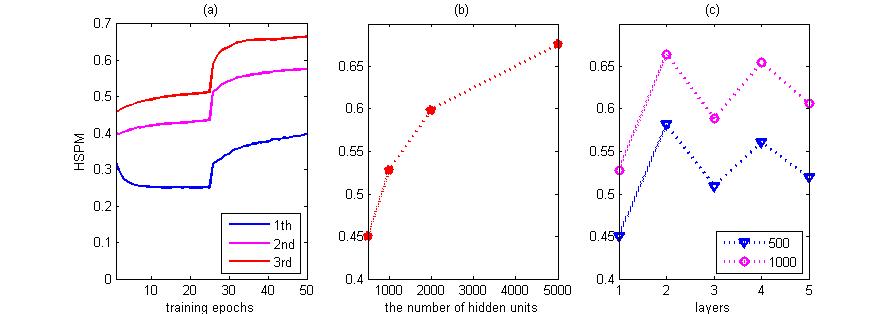}}
\vskip -0.15in
\caption{(\textbf{a-c}) Hoyer's sparseness measures (HSPM) of RBMs only pretrained on MNIST. (\textbf{a}) HSPM of three layers RBMs as the pretraining epoch increases (the momentum is 0.5 in the first 25 epochs and 0.9 in the rest 25 epochs). From down to top: RBMs from the 1st, 2nd and 3rd layers, respectively. (\textbf{b}) HSPM of RBMs with 500-5000 hidden units after 1000 training epochs. (\textbf{c}) HSPM of five layers RBMs with 500 and 1000 hidden units after 1000 training epochs.}
\label{fig:spon}
\end{center}
\vskip -0.2in
\end{figure*}

\begin{figure*}[ht]
\vskip -0.1in
\begin{center}
\centerline{\includegraphics[width=1.5\columnwidth]{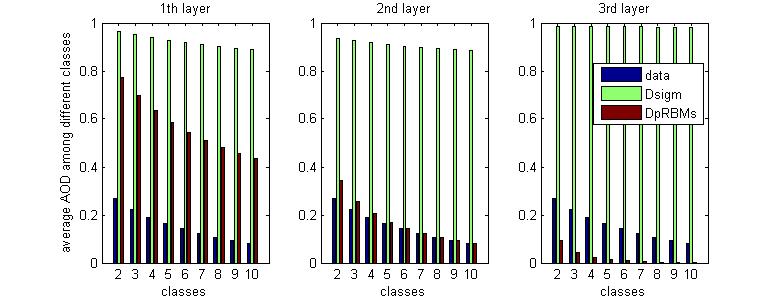}}
\vskip -0.15in
\caption{Activation overlapping degree (AOD) on MNIST. The left, middle and right respectively plot the average AOD among $k$ classes ($k$ changes from $2$ to $10$) in first, second and third layer.}
\label{fig:aodmnist}
\end{center}
\vskip -0.4in
\end{figure*}

Before presenting the comparison of activation overlapping units, we first show the sparseness of pretraining compared to the more traditional sigmoid activation function. The sparseness metric HSPM is the averaged value over the definition of \eqref{eq:hspm}.

We perform comparisons on MNIST, and results after fine-tuning training for 200 epochs are reported in Table~\ref{tab:hspm552}. The results show that compared to Dsigm the pretraining leads to models with higher sparseness, and smaller test errors. Table~\ref{tab:hspm552} compares the network HSPM of DBNs and DpRBMs to that of Dsigm. From Table~\ref{tab:hspm552}, we observe that the average sparseness of three layer DpRBMs is about $0.68$; the resulting DBNs has similar sparseness. In Fig~\ref{fig:spon}, (\textbf{a}) also shows that the feature of every layer RBMs is more sparse as the train epoch increases. In contract, the HSPM of Dsigm is on average below $0.14$.

When the pretraining are trained longer enough and the number of hidden unites increases, HSPM of the pretraining models will become more sparse and also has an upper bound. In Fig~\ref{fig:spon}, (\textbf{b}) shows that when the number of hidden units changes from 500 to 5000, an upper bounds of RBMs is 0.68 after 1000 training epochs.

As the number of layers increases, HSPM of the pretraining models also has an upper bound. From Fig~\ref{fig:spon}, (\textbf{c}) shows that upper bounds of five hidden (500 and 1000) layers RBMs are $0.58$ and $0.66$, respectively. We observe that the HSPM of the third layer pretraining is lower than one of the second layer. We empirically obtain the high HSPM by increasing the number of the hidden units of high layer, for example, DBNs (500-500-2000). This observation maybe can explain why the top layer should be big.

\begin{table}[!t]
\renewcommand{\arraystretch}{1.3}
\caption{Hoyer's sparseness measures (HSPM) of DNNs (50-100-100) on Birdsong.}
\label{tab:hspm511bird}
\vskip -0in
\centering
\small{
\begin{tabular}{c|c|c|c|c|c}
\hline\hline
         & dataset & 1st  & 2nd  & 3rd  & error  \\
\hline
DBNs   &  0.29   & 0.12 & 0.12 & 0.35 & 9.6$\%$  \\
DpRBMs &  0.29  & 0.62 & 0.68 & 0.59 &  \\
Dsigm  & 0.29    & 0.08 & 0.11 & 0.43 & 13.7$\%$ \\
\hline\hline
\end{tabular}}
\vskip -0.3in
\end{table}

We perform comparisons on Birdsong, results after fine-tuning training for 100 epochs are reported in Table~\ref{tab:hspm511bird}. The results also show that the pretraining leads to models with higher sparseness, and smaller test errors. From Table~\ref{tab:hspm511bird}, we observe that the sparseness of three layer DpRBMs is higher than that of Dsigm and database. Although after tuning the sparseness is close to Dsigm, the pretraining learn "good" initial values to initialize the DNN model parameters. This illustrates that the pretraining also is an optimization effect.

The HSPM in 3th layer are lower than 2nd layer. When training 2 layers networks, the resulting has similar test errors. So, there is a inspiration that the HSPM can be used to guide the number of layers and the number of hidden units.

\subsection{Comparison of Selective-Overlapping Units}
We perform comparisons on the test set of MNIST. For convenience, the test set $S$ is denoted by $\{\mathcal{X}_0,\cdots,\mathcal{X}_9\}$, where $\mathcal{X}_i (i=0,\cdots,9)$ represents a set of all digits $i$. $k$-combinations of the set $S$ is denoted by a set $S^k = \{S_1^k,\cdots,S_j^k,\cdots,S_{C_{10}^k}^k\}$, where $C_{10}^k$ is the number of $k$-combinations and $S_j^k$ is a subset of $k$ distinct elements of $S$. The average AOD among $k$ classes is an average of all AOD among a subset $S_j^k (j=1,\cdots, C_{10}^k)$.

We compare the average AOD of DpRBMs to that of Dsigm. We found that the pretraining can capture the characteristics \emph{(A1)} that there are many overlapping units in low layer and \emph{(A2)} that there are few (or no) overlapping units in high layer. From Fig~\ref{fig:aodmnist}, the results show that average AOD among $k$ classes ($k$ changes from $2$ to $10$) is high in low layer and is low (or zero) in high layer. The average AOD gets closer to $0$ as the number of layer increases in DpRBMs. Particularly, the average AOD gets closer to $0$ than data-self in the third layer. This reveals that it is easier to classify. But it is very approximate to $1$ in every layer of Dsigm.

\section{Conclusion}
\label{sec:cons}

Since the pretraining is known to perform well on MNIST, this paper mainly discusses why the unsupervised pretraining encourages moderate-sparseness. Our observations make us suspect that sparseness and activation overlapping degree play more important roles in deep neural networks. From Table~\ref{tab:hspm552}, Table~\ref{tab:hspm511bird} and Fig \ref{fig:aodmnist}, the pretraining can capture the sparse hidden units, the more activation overlapping units in low layer and the less (or no) activation overlapping units in high layers.

\nocite{langley00}
\small{
\bibliography{example_paper}
\bibliographystyle{icml2014}}

\end{document}